\documentclass[10pt,letterpaper]{article}
\usepackage[top=0.85in,left=1.0in,footskip=0.75in]{geometry}

\usepackage{url}
\makeatletter
\g@addto@macro{\UrlBreaks}{\UrlOrds}
\makeatother

\usepackage{changepage}

\usepackage[utf8]{inputenc}

\usepackage{textcomp,marvosym}

\usepackage{fixltx2e}

\usepackage{amsmath,amssymb}

\usepackage{cite}

\usepackage{nameref,hyperref}

\usepackage[right]{lineno}

\usepackage{rotating}

\usepackage{xcolor}

\raggedright
\usepackage[aboveskip=1pt,labelfont=bf,labelsep=period,justification=raggedright,singlelinecheck=off]{caption}

\makeatletter
\renewcommand{\@biblabel}[1]{\quad#1.}
\makeatother

\date{}

\newcommand{\acc}{$\mathit{Accuracy}$}
\newcommand{\favg}{$\overline{F_{1}}$}

\newcommand{\alfa}{$\mathit{Alpha}$}
\newcommand{\sn}{\hphantom{*}}
\newcommand{\se}{*}
\newcommand{\esr}{Emoji Sentiment Ranking}
\newcommand{\essr}{Emoji Sent. Rank.}
\newcommand{\sm}{\textsc{sem}}
\newcommand{\sd}{\textsc{sd}}
\newcommand{\cdf}{\textsc{cdf}}

\newcommand{\So}{\textbf{[So]}}

\begin{document}
\vspace*{0.35in}

\begin{flushleft}
{\Large
\textbf\newline{Sentiment of Emojis}
}
\newline
\\
Petra Kralj Novak,
Jasmina Smailovi\'{c},
Borut Sluban,
Igor Mozeti\v{c}
\\
\bigskip
Jo\v{z}ef Stefan Institute, Jamova 39, 1000 Ljubljana, Slovenia
\\
\bigskip

%
%





* Petra.Kralj.Novak@ijs.si, Igor.Mozetic@ijs.si

\end{flushleft}

\section*{Abstract} 
There is a new generation of emoticons, called emojis, that is increasingly
being used in mobile communications and social media.
In the past two years, over ten billion emojis were used on Twitter.
Emojis are Unicode graphic symbols, used as a shorthand to
express concepts and ideas. In contrast to the small number of well-known emoticons
that carry clear emotional contents, there are hundreds of emojis. 
But what are their emotional contents?
We provide the first emoji sentiment lexicon, called the Emoji Sentiment Ranking,
and draw a sentiment map of the 751 most frequently used emojis.
The sentiment of the emojis is computed from the sentiment
of the tweets in which they occur.
We engaged 83 human annotators to label over 1.6  million tweets in 13 European
languages by the sentiment polarity (negative, neutral, or positive).
About 4\% of the annotated tweets contain emojis.
The sentiment analysis of the emojis allows us to draw several interesting conclusions.
It turns out that most of the emojis are positive, especially the most popular ones.
The sentiment distribution of the tweets with and without emojis
is significantly different.
The inter-annotator agreement on the tweets with emojis is higher.
Emojis tend to occur at the end of the tweets, 
and their sentiment polarity increases with the distance.
We observe no significant differences in the emoji rankings
between the 13 languages and the Emoji Sentiment Ranking.
Consequently, we propose our
Emoji Sentiment Ranking as a European language-independent resource 
for automated sentiment analysis.
Finally, the paper provides a formalization of sentiment and
a novel visualization in the form of a sentiment bar.


\section{Introduction}

An \textbf{emoticon}, such as \texttt{;-)}, is shorthand for a facial expression.
It allows the author to express her/his feelings, moods and emotions,
and augments a written message with non-verbal elements.
It helps to draw the reader's attention, and enhances and improves
the understanding of the message.
An \textbf{emoji} is a step further, developed with modern communication
technologies that facilitate more expressive messages.
An emoji is a graphic symbol, ideogram, that represents not only facial
expressions, but also concepts and ideas, such as 
celebration, weather, vehicles and buildings, food and drink, animals and plants, or 
emotions, feelings, and activities.

Emojis on smartphones, in chat, and email applications have become 
extremely popular worldwide. 
For example, Instagram, an online mobile photo-sharing, 
video-sharing and social-networking platform, reported in March 2015 that 
nearly half of the texts on Instagram contained emojis \cite{Instagram}.
The use of emojis on the SwiftKey Android and iOS keybords,
for devices such as smartphones and tablets, was analyzed 
in the SwiftKey Emoji Report \cite{SwiftKey}, where a great variety
in the popularity of individual emojis, and even between countries, was reported.
However, to the best of our knowledge, no large-scale analysis of the
emotional content of emojis has been conducted so far.

Sentiment analysis is the field of study that analyzes people's opinions, 
sentiments, evaluations, attitudes, and emotions from a text 
\cite{liu2012,liu2015sentiment}.
In analyzing short informal texts, such as tweets, blogs or comments,
it turns out that the emoticons provide a crucial piece of information
\cite{boia2013worth,hogenboom2015exploiting,hogenboom2013exploiting,
davidov2010enhanced,liu2012emoticon,read2005using,zhao2012moodlens,
kiritchenko2014sentiment}. However, emojis have not been exploited so far,
and no resource with emoji sentiment information has been provided. 

In this paper we present the \esr, the first emoji sentiment lexicon of 751 emojis.
The lexicon was constructed from over 1.6 million tweets in 13 European languages,
annotated for sentiment by human annotators. 
In the corpus, probably the largest set of manually annotated tweets, 
4\% of the tweets contained emojis.
The sentiment of the emojis was computed from the sentiment of the tweets in which they occur,
and reflects the actual use of emojis in a context.

\paragraph*{Background.}

An emoticon is a short sequence of characters, typically punctuation symbols.
The use of emoticons can be traced back to the 19$^{th}$ century, 
when they were used in casual and humorous writing.
The first use of emoticons in the digital era is attributed to
professor Scott Fahlman, in a message on the computer-science message 
board of Carnegie Mellon University, on September 19, 1982.
In his message, Fahlman proposed to use \texttt{:-)} and
\texttt{:-(} to distinguish jokes from more serious posts.
Within a few months, the use of emoticons had spread,
and the set of emoticons was extended with hugs and kisses,
by using characters found on a typical keyboard.
A decade later, emoticons had found their way into everyday digital
communications and have now become a paralanguage of the web \cite{hogenboom2015exploiting}.

The word `emoji' literally means `picture character' in Japanese. 
Emojis emerged in Japan at the end of the 20$^{th}$ century to facilitate 
digital communication.
A number of  Japanese carriers (Softbank, KDDI, DoCoMo) provided their own
implementations, with incompatible encoding schemes. 
Emojis were first standardized in Unicode 6.0 \cite{Unicode6}---the 
core emoji set consisted of 722 characters. However,
Apple’s support for emojis on the iPhone, in 2010, led to global popularity.
An additional set of about 250 emojis was included in Unicode 7.0 \cite{Unicode7} in 2014.
As of August 2015, Unicode 8.0 \cite{Unicode8} defines a list of 1281 
single- or double-character emoji symbols.

\paragraph*{Related work.}

Sentiment analysis, or opinion mining, is the computational study of people's opinions, 
sentiments, emotions, and attitudes. It is one of the most active research areas 
in natural-language processing and is also extensively studied in data mining, 
web mining, and text mining \cite{liu2012,liu2015sentiment}.
The growing importance of sentiment analysis coincides with the growth of social media,
such as Twitter, Facebook, book reviews, forum discussions, blogs, etc.

The basis of many sentiment-analysis approaches is the sentiment lexicons, with the
words and phrases classified as conveying positive or negative sentiments.
Several general-purpose lexicons of subjectivity and sentiment have been constructed.
Most sentiment-analysis research focuses on English text and, consequently,
most of the resources developed (such as sentiment lexicons and corpora) are in English.
One such lexical resource, explicitly devised to support sentiment classification 
and opinion mining, is SentiWordNet 3.0\cite{baccianella2010sentiwordnet}. 
SentiWordNet extends the well-known WordNet \cite{Miller1995} 
by associating each synset with three numerical scores, describing 
how `objective', `positive', and `negative' the terms in the synset are.

Emoticons have proved crucial in the automated sentiment classification of informal texts 
\cite{boia2013worth,hogenboom2015exploiting,hogenboom2013exploiting,davidov2010enhanced,
liu2012emoticon,read2005using,zhao2012moodlens,kiritchenko2014sentiment}.
In an early work \cite{read2005using}, a basic distinction between positive 
and negative emoticons was used to automatically generate positive and
negative samples of texts. These samples were then used to train and test
sentiment-classification models using machine learning techniques.
The early results suggested that the sentiment conveyed by emoticons 
is both domain and topic independent.
In later work, these findings were applied to automatically construct 
sets of positive and negative tweets \cite{go2009,davidov2010enhanced,pak2010twitter},
and sets of tweets with alternative sentiment categories, 
such as the angry and sad emotional states \cite{zhao2012moodlens}.
Such emoticon-labeled sets are then used to automatically train the sentiment classifiers.
Emoticons can also be exploited to extend the more common features used
in text mining, such as sentiment-carrying words.
A small set of emoticons has already been used as additional features 
for polarity classification \cite{davidov2010enhanced,thelwall2010sentiment}.
A sentiment-analysis framework that takes explicitly into account the information conveyed 
by emoticons is proposed in \cite{hogenboom2015exploiting}.

There is also research that analyzes graphical emoticons and their sentiment,
or employs them in a sentiment classification task.
The authors in~\cite{amalanathan2015social} manually
mapped the emoticons from Unicode 8.0 to nine emotional
categories and performed the sentiment classification of tweets, using both
emoticons and bag-of-words as features. 
Ganesan et al.~\cite{ganesan2008mining} presents a system for adding the
graphical emoticons to text as an illustration of the written emotions.

Several studies have analyzed emotional contagion through posts on
Facebook and showed that the emotions in the posts of online friends 
influence the emotions expressed in newly generated 
content \cite{kramer2012spread,kramer2014experimental,coviello2014detecting,Zollo2015}.
Gruzd et al.~\cite{gruzd2011happiness} examined the spreading of emotional
content on Twitter and found that the positive posts are
retweeted more often than the negative ones.
It would be interesting to examine how the presence of emojis in
tweets affects the spread of emotions on Twitter, i.e., to relate
our study to the field of emotional contagion~\cite{hatfield1994emotional}.

\paragraph{Contributions.}
Emojis, a new generation of emoticons, are increasingly being used in social media.
Tweets, blogs and comments are analyzed to estimate the emotional attitude
of a large fraction of the population to various issues.
An emoji sentiment lexicon, provided as a result of this study, is a valuable
resource for automated sentiment analysis. 
The \esr\, has a format similar to SentiWordNet \cite{baccianella2010sentiwordnet}, 
a publicly available resource for opinion mining, used in more than 700 
applications and studies so far, according to Google Scholar.
In addition to a public resource, the paper provides an in-depth analysis
of several aspects of emoji sentiment.
We draw a sentiment map of the 751 emojis, compare the differences between
the tweets with and without emojis, the differences between the more and less
frequent emojis, their positions in tweets, and the differences between
their use in the 13 languages. Finally, a formalization of sentiment
and a novel visualization in the form of a sentiment bar are presented.

\section{Results and Discussion}

\subsection{Emoji sentiment lexicon}

The sentiment of emojis is computed from the sentiment of tweets.
A large pool of tweets, in 13 European languages, was labeled for
sentiment by 83 native speakers. Sentiment labels can take one of three
ordered values:
\textit{negative} $\prec$ \textit{neutral} $\prec$ \textit{positive}.
A sentiment label, $c$, is formally a discrete, 3-valued variable $c \in \{-1, 0, +1\}$.
An emoji is assigned a sentiment from all the tweets in which it occurs.
First, for each emoji, we form a discrete probability distribution
($p_{-}$, $p_{0}$, $p_{+}$).
The sentiment score $\overline{s}$ of the emoji is then computed as 
the mean of the distribution.
The components of the distribution, i.e., $p_{-}$, $p_{0}$, and $p_{+}$ denote the
negativity, neutrality, and positivity of the emoji, respectively.
The probability $p_c$ is estimated from the number of occurrences, $N$, of the
emoji in tweets with the label $c$. Note that an emoji can occur multiple times
in a single tweet, and we count all the occurrences.
A more detailed formalization of the sentiment representation can be found in
the Methods section.

\begin{figure}[h]
\begin{center}
\includegraphics[width=13cm]{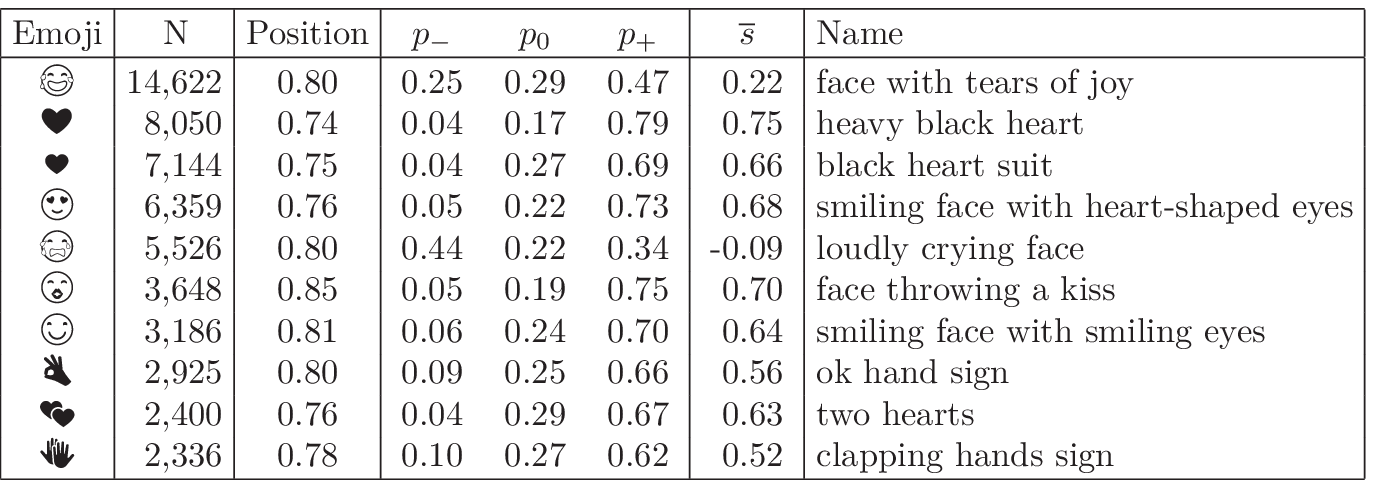}
\caption{\textbf{Top 10 emojis.}
Emojis are ordered by the number of occurrences $N$.
The average \textit{position} ranges from 0 (the beginning of the tweets) to 1 (the end of the tweets).
$p_c$, $c \in \{-1, 0, +1\}$, are the negativity, neutrality, and positivity, respectively.
$\overline{s}$ is the sentiment score.}
\label{fig:sentiment-tab}
\end{center}
\end{figure}

We thus form a sentiment lexicon of the 751 most frequent emojis,
called the \esr. The complete \esr\, is available as a web page at
\url{http://kt.ijs.si/data/Emoji_sentiment_ranking/}.
The 10 most frequently used emojis from the lexicon are shown 
in Fig~\ref{fig:sentiment-tab}.

First we address the question of whether the emojis in our lexicon are representative.
We checked \textbf{Emojitracker} (\url{http://emojitracker.com/}), a website
that monitors the use of emojis on Twitter in realtime.
In the past two years, Emojitracker has detected almost 10 billion emojis
on Twitter! From the ratio of the number of emoji occurrences and tweets in our 
dataset ($\sim 2.3$), we estimate that there were about 4~billion tweets with emojis.
In our dataset of about 70,000 tweets, we found 969 different emojis, 721 of them in 
common with Emojitracker.

We compared the emojis in both sets, ordered by the number of occurrences,
using Pearson's \cite{Pearson1895} and Spearman's rank \cite{Spearman1904}
correlation. We successively shorten our list of emojis by cutting off
the least-frequent emojis. The results for two thresholds, $N \ge 1$ and $5$,
with the highest correlation coefficients, are shown in Table~\ref{tab:emojitracker}. 
Both correlation coefficients are high, significant at the 1\% level, thus
confirming that our list of emojis is indeed representative of their general
use on Twitter. Between the two options, we decided to select the list
of emojis with at least $5$ occurrences, resulting in the lexicon of 751 emojis.
The sentiment scores for the emojis with fewer then $5$ occurrences are not very reliable.

\begin{table}[h]
  \centering
  \caption{\textbf{Overlap with Emojitracker.}
  Correlations are between the occurrences of emojis in the \esr\, and Emojitracker, 
  for two minimum occurrence thresholds.
  The numbers in parenthesis are the emojis that are common in both sets.
  The correlation values, significant at the 1\% level, are indicated by *.}
    \begin{tabular}{|l|r|c|c|c|c|}
    \hline
                & Tweets\sn\sn   & Different    & Pearson       & Spearman rank \\
            & with emojis  & emojis used  & correlation   & correlation   \\ 
    \hline
    \textbf{Emojitracker}  &  $\sim$4 billion    & 845    & /        & / \\
    \hline
    \essr & & & & \\
    $N \ge 1$          &  69,673    & 969 (721)       & 0.945\se & 0.897\se \\
    \essr & & & & \\
    $N \ge 5$          &  69,546    & \textbf{751} (608)       & 0.944\se & 0.898\se \\
    \hline
    \end{tabular}
  \label{tab:emojitracker}
\end{table}

\newpage
\subsection{Emoji sentiment map}

\begin{figure}[ht]
\begin{center}
\includegraphics[width=\textwidth]{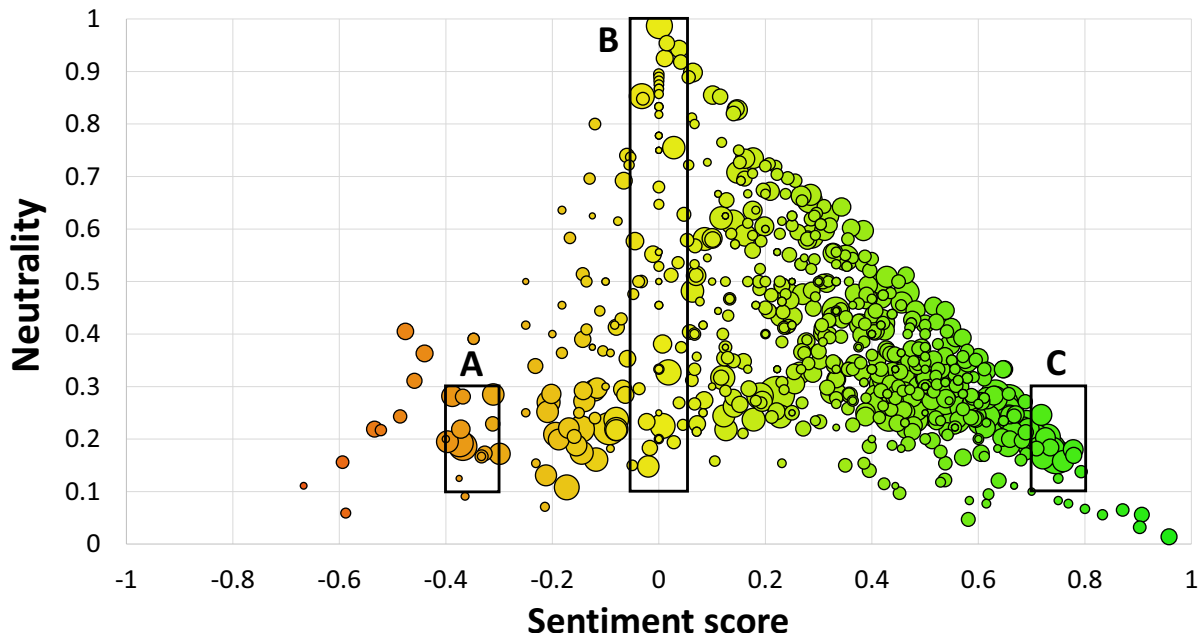}
\caption{\textbf{Sentiment map of the 751 emojis.}
Left: negative (red), right: positive (green), top: neutral (yellow). 
Bubble size is proportional to $\log_{10}$ of the emoji occurrences in the \esr. 
Sections A, B, and C are references to the zoomed-in panels in Fig~\ref{fig:zoom}.}
\label{fig:sentiment-map}
\end{center}
\end{figure}

Before we analyze the properties of the tweets with emojis, we first discuss
two visualizations of the \esr. 
Fig~\ref{fig:sentiment-map} shows the overall map of the 751 emojis.
The position of an emoji is determined by its sentiment score $\overline{s}$
and its neutrality $p_0$. The sentiment score $\overline{s}$ is in the range
$(-1, +1)$ and is computed as $p_{+} - p_{-}$. The more positive emojis
are on the right-hand side of the map (green), while the negative ones are
on the left-hand side (red). The neutral emojis (yellow) span a whole band around 
$\overline{s} = 0$. The emojis are prevailingly positive, the mean sentiment score
is $+0.3$ (see 
Fig~\ref{fig:emoji-distribution}).
The bubble sizes are proportional to the number of occurrences.

\begin{figure*}[h]
\begin{center}
\includegraphics[width=\textwidth]{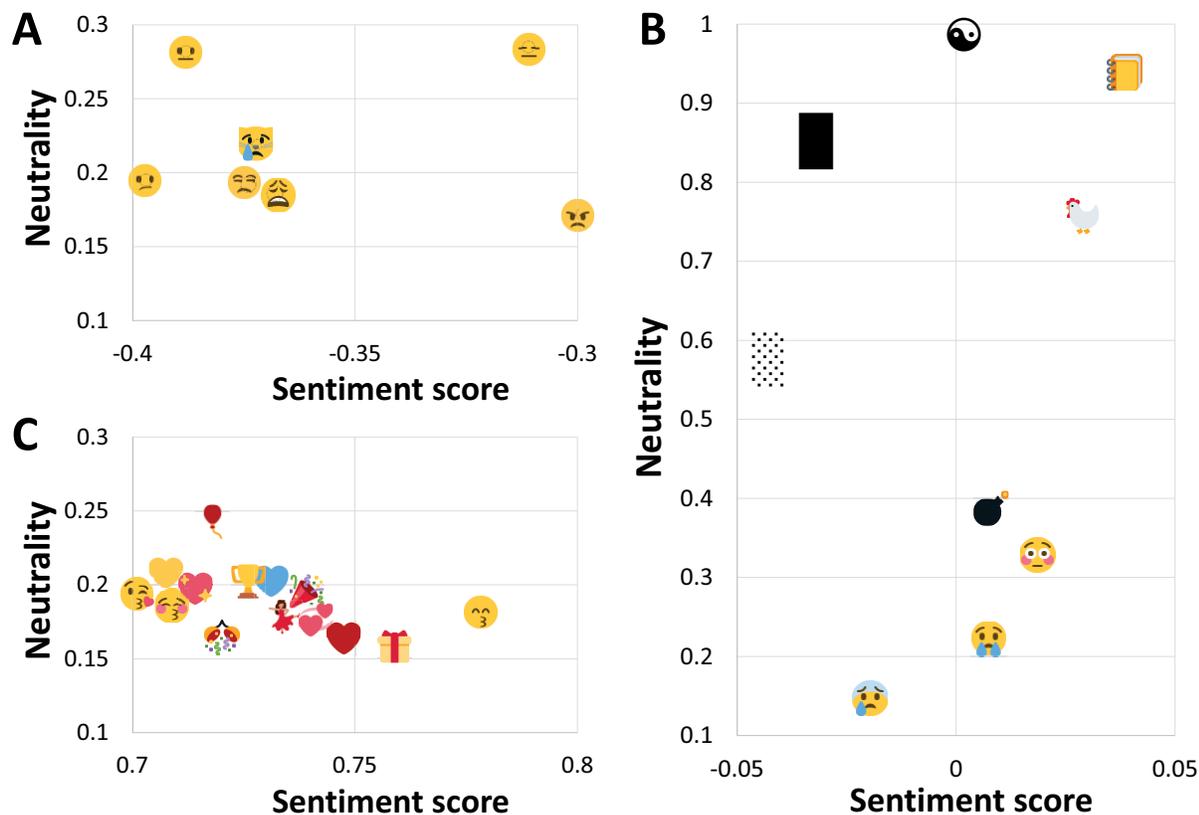}
\caption{\textbf{Emojis in sections A, B, and C of Fig~\ref{fig:sentiment-map}.} 
Shown are emojis that occur at least 100 times in the \esr. 
Panel {\textbf A}: negative emojis, panel {\textbf B}: neutral (top) and 
bipolar (bottom) emojis, panel {\textbf C}: positive emojis.}
\label{fig:zoom}
\end{center}
\end{figure*}

A more detailed view of some actual emojis on the map is shown in Fig~\ref{fig:zoom}.
The most frequent negative emojis (panel A) are sad faces.
On the other hand, the most frequent positive emojis (panel C) are not only happy
faces, but also hearts, party symbols, a wrapped present, and a trophy.
Even more interesting are the neutral emojis (panel B).
All of them have a sentiment score around $0$, but the neutrality
$p_{0}$ ranges between $0$ and $1$. The bottom two, with low $p_{0}$
(face with cold sweat, crying face), are bipolar, with a high negativity and positivity,
where $p_{-} \approx p_{+}$.
The middle two (flushed face, bomb) have a uniform sentiment distribution,
where $p_{-} \approx p_{0} \approx p_{+}$.
The top ones, with high $p_{0}$, are neutral indeed, symbolized by the
yin yang symbol at the very top.

\newpage
\subsection{Tweets with and without emojis}

In this subsection we analyze the interplay of the human perception of
tweets that are with and without emojis. If we consider the sentiment of a tweet
as a rough approximation of its emotional content, we can ask two questions.
Are the tweets with emojis more emotionally loaded? 
Does the presence of emojis in tweets have an impact on the human 
emotional perception of the tweets?
We do not draw any causal conclusions, but report the results of two
experiments that indicate that the answer to both questions is positive.

First, we compare all the manually labeled tweets that are with and without emojis.
From the distribution of the negative, neutral, and positive tweets in
both sets, we compute the mean, standard deviation (\sd), and
standard error of the mean (\sm). The results are shown 
in Table~\ref{tab:label_distribution_entire_and_emoji_dataset}.

\begin{table}[h]
\centering
\caption{\textbf{Sentiment of tweets with and without emojis.}
For each set, the mean, \sd\, and \sm\, are computed from the distribution of
negative, neutral, and positive tweets.}
\label{tab:label_distribution_entire_and_emoji_dataset}
\begin{tabular}{| l | c | c |}
 \hline
                    & \textbf{Tweets}          & \textbf{Tweets} \\
  Sentiment         & \textbf{with emojis}     & \textbf{without emojis}  \\ 
  \hline
  \textit{Negative}          & 12,156 (17.5\%)   & 410,301 (26.1\%)  \\
  \textit{Neutral}           & 19,938 (28.6\%)   & 587,337 (37.3\%)  \\
  \textit{Positive}          & 37,579 (53.9\%)   & 576,424 (36.6\%)  \\
  \hline
  Total             & 69,673            & 1,574,062         \\ 
  \hline
  Mean              & +0.365            & +0.106 \\
  \sd, \sm          & 0.762, 0.0029     & 0.785, 0.0006 \\ 
  \hline
  \end{tabular}
\end{table}

We test the null hypothesis that the two populations have equal means.
We apply Welch's t-test \cite{Welch1947} for two samples with unequal
variances and sizes. 
We are aware that the two populations might not be normally distributed,
but Welch's t-test is robust for skewed distributions, and even more so
for large sample sizes \cite{Fagerland2012}.
With $t = 87$, the degrees of freedom $\gg$ 100
(due to large sample sizes), and the p-value $\approx 0$, 
the null hypothesis can be rejected.
We can conclude, with high confidence, that the tweets with and without emojis
have significantly different sentiment means.
Additionally, the tweets with emojis are significantly more positive 
(mean $= +0.365$) than the tweets without emojis (mean $= +0.106$).

Next, we compare the agreement of the human annotators on the tweets with and 
without emojis. The Twitter sentiment classification is not an easy task
and humans often disagree on the sentiment labels of controversial tweets.
During the process of annotating the 1.6 million tweets, we found 
that even individual annotators are not consistent with themselves.
Therefore, we systematically distributed a fraction of the tweets to
be annotated twice in order to estimate the level of agreement.
This annotator self-agreement is a good indicator of the reliability of
the annotator. The inter-annotator agreement, on the other hand,
indicates the difficulty of the task. In the case of emojis, our goal
is to verify whether their presence in tweets correlates with 
a higher inter-annotator agreement.

There are a number of measures to estimate the inter-annotator agreement.
We apply three of them from two different fields, to ensure robust estimates.
The first one, Krippendorff's \alfa-reliability \cite{Krippendorff2012},
generalizes several specialized agreement measures. 
When the annotators are in perfect agreement, \alfa\;$=1$, and when the level of 
agreement equals the agreement by chance, \alfa\;$=0$. 
We applied an instance of \alfa\, that takes into account the ordering of labels
and assigns a higher penalty to more extreme disagreements.
For example, a disagreement between the
\textit{negative} and the \textit{positive} sentiment is four times as costly
as that between the \textit{neutral} and \textit{positive}.

The simplest measure of agreement is the joint probability of agreement,
also known as \acc\,, when evaluating classification models.
\acc\, is the number of equally labeled tweets by different annotators,
divided by the total number of tweets labeled twice.
It assumes the data labels are unordered (nominal) and does not take into 
account the agreement by chance, but it is easy to interpret.

The third measure comes from the field of machine learning.
It is used to evaluate the performance of classification models against 
a test set, where the true sentiment label is known.
The measure, \favg($-,+$), is a standard 
measure of performance, specifically designed for a 3-valued
sentiment classification \cite{kiritchenko2014sentiment},
where the \textit{negative} ($-$) and \textit{positive} ($+$) sentiments 
are considered more important than the \textit{neutral} one.
Here, we adapt it to estimate the agreement of a pair of annotators.

\begin{table}[h]
\centering
\caption{\textbf{Inter-annotator agreement on tweets with and without emojis.}
The agreement is computed in terms of three measures over a subset of tweets that
were labeled by two different annotators.}
\label{tab:inter_annotator_agreement}
\begin{tabular}{| l | c | c |}
 \hline
  Agreement       & \textbf{Tweets}      & \textbf{Tweets} \\
  measure         & \textbf{with emojis} & \textbf{without emojis}  \\
  \hline
  \alfa           & 0.597       & 0.495 \\
  \acc            & 0.641       & 0.583 \\
  \favg($-,+$)    & 0.698       & 0.598 \\
  \hline
  No. of tweets   & & \\
  annotated twice & 3,547 & 52,027 \\
  \hline
\end{tabular}
\end{table}

Table~\ref{tab:inter_annotator_agreement} gives the results of the
inter-annotator agreements on the tweets with and without emojis.
Coincidence matrices for both cases are in the Methods section,
in Tables~\ref{tab:inter_emojis} and \ref{tab:inter_noemojis}, respectively.
All three measures of agreement, \alfa, \acc, and \favg($-,+$),
are considerably higher for the tweets with emojis, 
by 21\%, 10\%, and 17\%, respectively.
We do not give any statistical-significance results, but it seems
safe to conclude that the presence of emojis has a positive impact
on the emotional perception of the tweets by humans.
After all, this is probably the main reason why they are used
in the first place.

\subsection{Sentiment distribution}

\begin{figure}[h]
\begin{center}
\includegraphics[width=13cm]{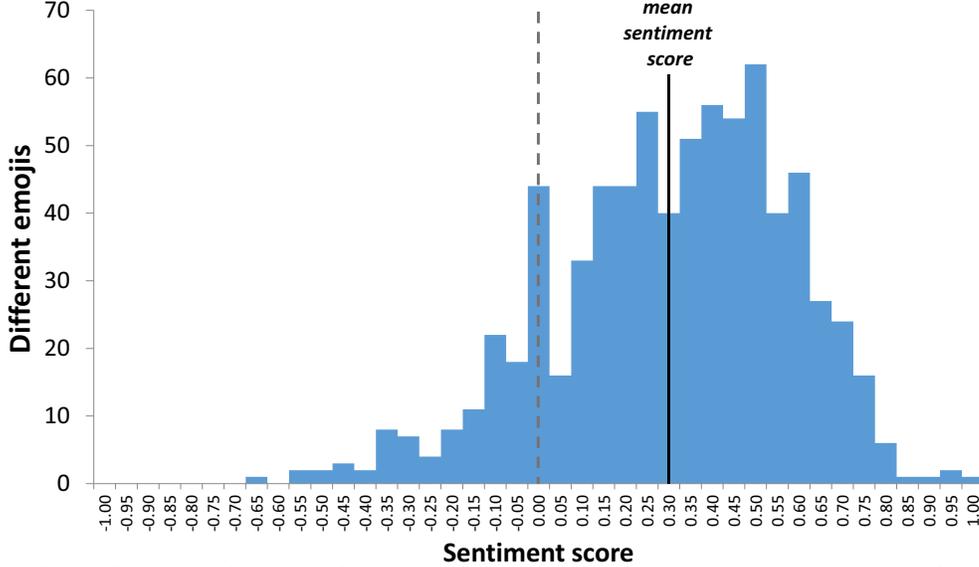}
\caption{\textbf{Distribution of emojis by sentiment score.}
The mean sentiment score of the 751 emojis (in bins of $0.05$) is +0.305.}
\label{fig:emoji-distribution}
\end{center}
\end{figure}

In this subsection we analyze the sentiment distribution of the emojis
with respect to the frequency of their use.
The question we address is the following:
Are the more frequently used emojis more emotionally loaded?
First, in Fig~\ref{fig:emoji-distribution} we show the sentiment distribution
of the 751 emojis, regardless of their frequencies.
It is evident that the sentiment score of the emojis is approximately 
normally distributed, with mean = $+0.3$, prevailingly positive.

\begin{figure}[h]
\begin{center}
\includegraphics[width=\textwidth]{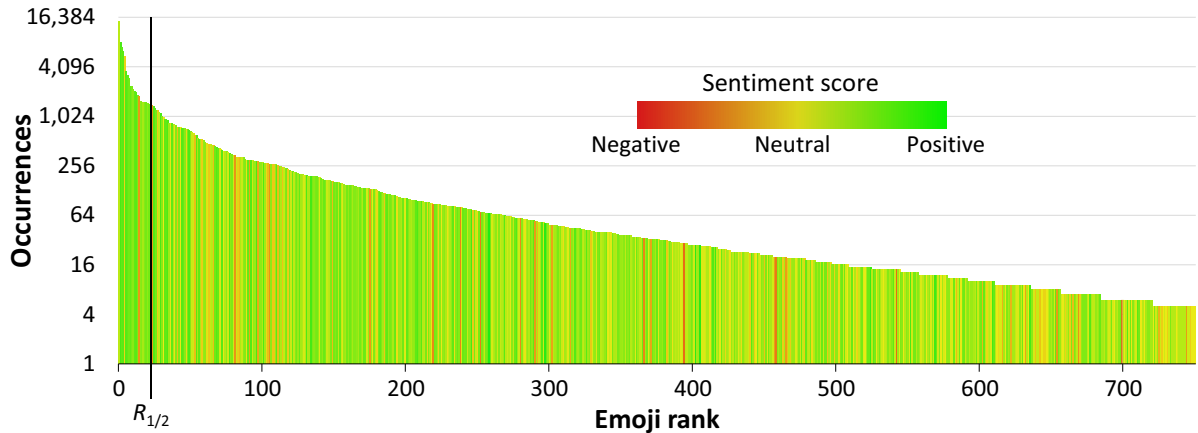}
\caption{\textbf{Distribution of occurrences and sentiment of the 751 emojis.}
The emojis are ranked by their occurrence (log scale).
The column color indicates the sentiment score. The partitioning into two
equally weighted halfs is indicated by a line at $R_{1/2}$.
The first 33 emojis are zoomed-in in Fig~\ref{fig:33volume}.}
\label{fig:SRvolumes}
\end{center}
\end{figure}

\begin{figure}[h]
\begin{center}
\includegraphics[width=\textwidth]{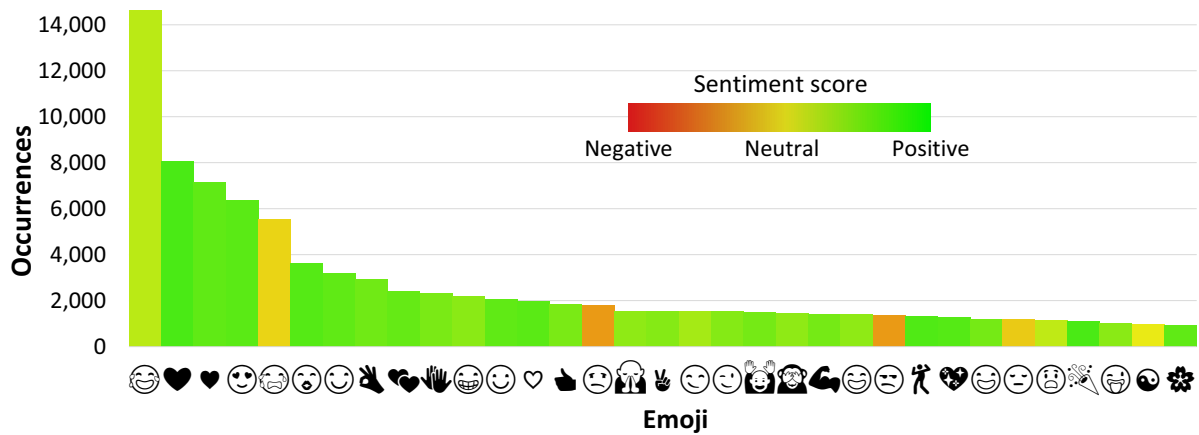}
\caption{\textbf{Top 33 emojis by occurrence.}
Column color represents the emoji sentiment score.}
\label{fig:33volume}
\end{center}
\end{figure}

In Fig~\ref{fig:SRvolumes} we rank the emojis by the number of their
occurrences in tweets. 
The sentiment score of each emoji is indicated
by the color. The zoomed-in section of the first 33 emojis is
in Fig~\ref{fig:33volume}.

We did not thoroughly analyze the frequency-rank distribution
of the emojis. A quick analysis suggests that the data follows a
power law with an exponential cutoff at a rank of about $200$.
Using a maximum-likelihood estimator \cite{Newman2005}, the exponent 
of the power law is estimated to be $-1.3$, a relatively extreme exponent.
Even more relevant is the distribution of the emojis on Emojitracker, but
this remains a subject of further research. Here we concentrate on
the sentiment distribution.

We define a cumulative distribution function $\cdf(R)$ of rank $R$
over a set of ranked emojis as:
$$
\cdf(R) = N(r \le R) = \sum_{r \le R} N(r) \; ,
$$
where $r$ denotes the rank of an emoji, and $N(r)$ the number of
occurrences of the emoji at rank $r$.
In order to compare the higher-ranked emojis (more frequent)
with the lower-ranked ones (less frequent), we define a midpoint rank
$R_{1/2}$, such that:
$$
N(1 \le r \le R_{1/2}) \approx N(R_{1/2} < r \le 751) \; .
$$
The midpoint rank $R_{1/2}$ partitions the emojis into two subsets
with an approximately equal cumulative number of occurrences. In the case
of the \esr, the midpoint is at $R_{1/2} = 23$.

We compute the mean sentiment, \sd, and \sm\, of the more frequent and
the less frequent emojis. 
The results are shown in Table~\ref{tab:first_vs_second_half}.

\begin{table}[h]
\centering
\caption{\textbf{Comparison of the more-frequent with the less-frequent emojis.}
The emojis ($r$) ranked by occurrence $N(r)$ are partitioned into two halves
with approximately the same cumulative number of occurrences.}
\label{tab:first_vs_second_half}
\begin{tabular}{|l|cc|c|}
 \hline
  & \textbf{1st half} $(r \le 23)$ & \textbf{2nd half} $(23 < r)$ & Total \\
 \hline
 Different emojis  & 23         & 728           & 751 \\
 Occurrences $(\sum N(r))$    & 77,969        & 78,488        & 156,457 \\
 \hline
 Sentiment mean & +0.463        & +0.311        & +0.387 \\
 \sd, \sm       & 0.280, 0.0010 & 0.319, 0.0011 & 0.300, 0.0008 \\
 \hline
\end{tabular}
\end{table}

We test the null hypothesis that the two populations of emojis have
equal mean sentiment scores. Again, we apply Welch's t-test for two
samples with unequal variances, but similar sizes. With $t = 100$,
the degrees of freedom $\gg 100$ (due to large sample sizes), and
the p-value $\approx 0$, the null hypothesis can be rejected.
We can conclude, with high confidence, that the more-frequent
emojis are significantly more positive than the less-frequent ones.

This result supports the thesis that the emojis that are used more often
are more emotionally loaded, but we cannot draw any causal conclusion.
Are they more positive because they are more often used in positive
tweets, or are they more frequently used, because they are more positive?

\newpage
\subsection{Sentiment and emoji position}

\begin{figure}[h]
\begin{center}
\includegraphics[width=13cm]{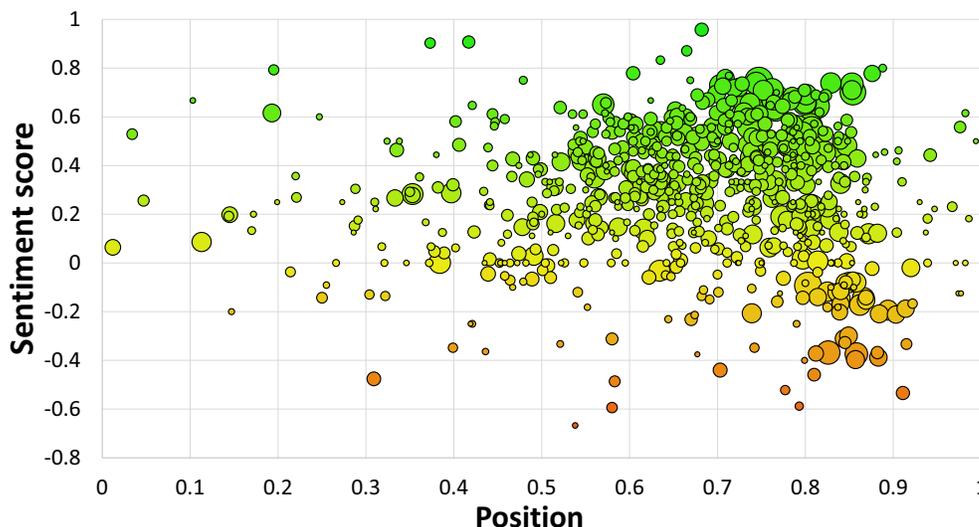}
\caption{\textbf{Average positions of the 751 emojis in tweets.} 
Bubble size is proportional to $\log_{10}$ of the emoji occurrences in the \esr.
Left: the beginning of tweets, right: the end of tweets, 
bottom: negative (red), top: positive (green).}
\label{fig:Emoji_position}
\end{center}
\end{figure}

Where are the emojis typically placed in tweets?
Emoticons such as \textsc{:-)} are used sparsely and typically at
the very end of a sentence. Emojis, on the other hand, appear in groups
and not only at the end of the tweets.
Fig~\ref{fig:Emoji_position} shows the average positions of the 751 emojis
in the tweets. On average, an emoji is placed at $2/3$ of the length of a tweet.

\begin{figure}[h]
\begin{center}
\includegraphics[width=13cm]{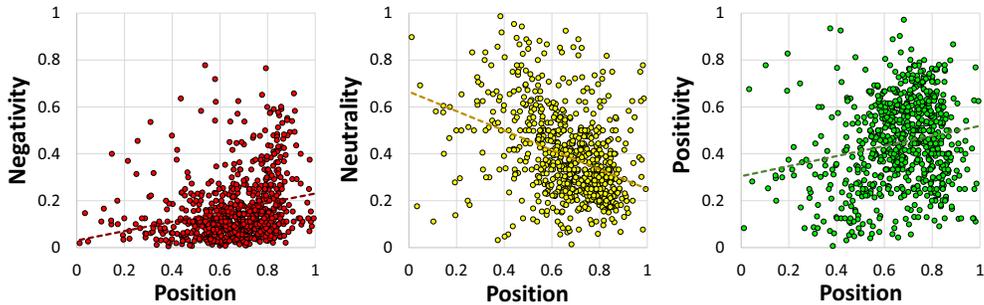}
\caption{\textbf{Negativity, neutrality, and positivity regressed with position
(from left to right).} The trendlines are functions $p_{c}(d)$
of the distance $d$ from the beginning of the tweets.}
\label{fig:nnp_position}
\end{center}
\end{figure}

Fig~\ref{fig:Emoji_position} also indicates the sentiment of an emoji
in relation to its position. In Fig~\ref{fig:nnp_position} we decompose
the sentiment into its three constituent components and show the regression
trendlines.

The linear regression functions in Fig~\ref{fig:nnp_position} have the following
forms:
$$
\textit{negativity:} \;\;\; p_{-}(d) = 0.20d + 0.03 \;\; (R^2 = 0.06)\;,
$$
$$
\textit{neutrality:} \;\;\; p_{0}(d) = -0.41d + 0.66 \;\; (R^2 = 0.14)\;,
$$
$$
\textit{positivity:} \;\;\; p_{+}(d) = 0.21d + 0.30 \;\; (R^2 = 0.04)\;,
$$
where $d$ is the distance from the beginning of the tweets.
The functions do not fit the data very well, but they give some useful insight.
At any distance $d$, and for any subset of emojis, the component probabilities add up to $1$:
$$
\sum_{c}p_c(d) = 1 
$$
However, the negativity and positivity increase with the distance, whereas 
the neutrality decreases. This means that more emotionally loaded emojis,
either negative or positive, tend to occur towards the end of the tweets.

\subsection{Emojis in different languages}

In the final subsection we analyze the use of emojis in the 13 languages
processed in this study. Can the \esr\, be considered a universal resource,
at least for European languages? Is the sentiment ranking between the
different languages significantly different? The results in Table~\ref{tab-lang}
indicate that the answer to the first question is positive and that there
is no evidence of significant differences between the languages.

\begin{table}[h]
  \centering
  \caption{\textbf{Emoji sentiment in different languages.}
  The languages are ordered by the number of different emojis used.
  Correlations are between the sentiment scores of emojis in the 13 languages
  and the \esr.
  The correlation values, significant at the 1\% level, are indicated by *.}
    \begin{tabular}{|l|r|c|c|c|c|}
    \hline
            & Tweets\sn\sn   & Different    & Pearson       & Spearman rank \\
            & with emojis  & emojis used  & correlation   & correlation   \\ 
    \hline
    \textbf{\essr}  &  69,546 \sn              & 751               & / & / \\
    \hline
    English         &  19,819 \sn              & 511               & 0.834\se & 0.819\se \\
    Spanish         &  22,063 \sn              & 448               & 0.552\se & 0.573\se \\
    Polish          &  8,112  \sn              & 253               & 0.810\se & 0.783\se \\
    Russian         &  5,007  \sn              & 221               & 0.777\se & 0.756\se \\
    Hungarian       &  2,324  \sn              & 176               & 0.588\se & 0.612\se \\
    German          &  3,062  \sn              & 142               & 0.782\se & 0.783\se \\
    Swedish         &  2,797  \sn              & 139               & 0.702\se & 0.674\se \\
    Ser/Cro/Bos     &  2,096  \sn              & 123               & 0.708\se & 0.615\se \\
    Slovak          &  1,526  \sn              & 108               & 0.620\se & 0.499\se \\
    Slovenian       &    996  \sn              & \sn66             & 0.526\se & 0.541\se \\
    Portuguese      &    796  \sn              & \sn56             & 0.410\se & 0.429\se \\
    Bulgarian       &    607  \sn              & \sn36             & 0.557\se & 0.443\se \\
    Albanian        &    341  \sn              & \sn19             & 0.363\sn & 0.416\sn \\ 
    \hline
    \end{tabular}
  \label{tab-lang}
\end{table}%

For each language, we form a list of emojis used in the collected tweets of the language,
cut off the emojis with fewer than $5$ occurrences (the same threshold as applied to 
the overall \esr), and compute their sentiment score.
We compute the correlation coefficients between the \esr\, and the individual languages.
As can be seen in Table~\ref{tab-lang}, the number of emojis actually used in the different
languages (above the threshold) drops considerably.
However, their sentiment scores and ranking remain stable.
Both Pearson's correlation and Spearman's rank correlation are relatively high,
and significant for all the languages, except Albanian.
This result is biased towards languages with more tweets since they have
a larger share in the joint \esr. An alternative test might compare individual
languages and the \esr\, with the language removed.
However, as a first approximation, it seems
reasonable to use the \esr\, as a universal, language-independent resource, at 
least for European languages.

\section{Conclusions}

In this paper we describe the construction of an emoji sentiment lexicon,
the \esr, the first such publicly available resource.
We have formalized and analyzed the sentiment properties of the emojis in depth
and highlighted some interesting conclusions.

The data that enabled these analyses, 1.6 million annotated tweets in 
13 different languages, is a valuable resource with many other useful applications.
In particular, we are constructing sentiment-classification models
for different languages, and applying them to various tasks.
The Slovenian and Bulgarian language-sentiment models were already applied
to monitor the mood on Twitter during political elections
in realtime \cite{smailovic2015}.
The English sentiment model was used to compare the sentiment leanings
of different retweet network communities towards various
environmental topics \cite{Sluban2015}.
A domain-specific English sentiment model (from another set of financial tweets)
was applied to analyze the effects of Twitter sentiment on 
stock prices \cite{Ranco2015}.
Yet another English sentiment model was constructed
by combining a large set of general, emoticon-labeled tweets with 
domain-specific financial tweets, and tested for Granger causality
on the Baidu stocks \cite{smailovic2014}.
The same methodology of manual text annotations, automated
model construction, and sentiment classification was also applied 
to Facebook comments in Italian, where the emotional dynamics in the spreading
of conspiracy theories was studied \cite{Zollo2015}. 

The sentiment annotation of tweets by humans is expensive.
Emoticons were already used as a proxy for the sentiment labels of tweets.
We expect that the \esr\, will turn out to be a valuable resource for
helping humans during the annotation process, or even to 
automatically label the tweets with emojis for sentiment.
In a lexicon-based approach to sentiment analysis, the emoji lexicon
can be used in combination with a lexicon of sentiment-bearing words.
Alternatively, an emoji with already-known sentiment can act as
a seed to transfer the sentiment to the words in proximity.  
Such a corpus-based approach can be used for an automated corpus
construction for feature generation \cite{kiritchenko2014sentiment},
and then applied to train a sentiment classifier.

There are other dimensions of sentiment that are beyond the one-dimensional
scale from negativity to positivity and worth exploring.
The expressiveness of the emojis allows us to assign them more subtle
emotional aspects, such as anger, happiness, or sadness, and some
shallow semantics, such as activities, locations, or objects of interest.
An additional structuring of the emojis can be derived from correlations
between their sentiment, e.g., various versions of hearts expressing love.
However, we consider the interplay between the emojis and the text to be one
of the most promising directions for future work. Not only the position 
of an emoji, but certainly its textual context is also important
in determining the role of the emoji as an amplifier and modifier of
the meaning.

In the future, it will be interesting to monitor how the use of emojis is growing,
and if textual communication is increasingly being replaced
by a pictorial language.
Also, the sentiment and meaning of emojis evolve over time.
It might be interesting to investigate the convergence of agreement on the
meaning of controversial emojis, and to study the underpinnings of the
corresponding social processes.

\section{Methods}

\paragraph*{Ethics statement.}

The tweets were collected through the public Twitter API and 
are subject to the Twitter terms and conditions.
The sentiment annotations were supported by the Goldfinch platform,
provided by Sowa Labs (\url{http://www.sowalabs.com}).
The human annotators were engaged for the purpose, and were aware 
that their annotations will be used to construct the 
sentiment-classification models, and to estimate the 
inter-annotator agreement and the annotator self-agreement.

\subsection{Data collection}

The main source of the data used in this study is a collection of tweets,
in 13 European languages, collected between April 2013 and February 2015.
Most of the tweets (except English) were collected during a joint
project with Gama System (\url{http://www.gama-system.si}), 
using their PerceptionAnalytics platform 
(\url{http://www.perceptionanalytics.net}).
The tweets of selected languages were collected through Twitter Search API,
by specifying the geolocations of the largest cities.
For English tweets, we used Twitter Streaming API (a random sample of
1\% of all public tweets), and filtered out the English posts.

\begin{table}[h]
  \centering
  \caption{\textbf{Tweets annotated for sentiment in different languages.}
  Languages are in alphabetical order, Ser/Cro/Bos denotes a union of tweets in Serbian,
  Croatian and Bosnian.}
  \begin{tabular}{|l|r|r|}
\hline
            & No. of  & No. of \\
\textbf{Language}    & tweets  & annotators \\
\hline
Albanian    &  53,005 & 13 \\
Bulgarian   &  67,169 & 18 \\
English     & 103,034 &  9 \\
German      & 109,130 &  5 \\
Hungarian   &  68,505 &  1 \\
Polish      & 223,574 &  8 \\
Portuguese  & 157,393 &  1 \\
Russian     & 107,773 &  1 \\
Ser/Cro/Bos & 215,657 & 13 \\
Slovak      &  70,425 &  1 \\
Slovenian   & 133,935 &  7 \\
Spanish     & 275,588 &  5 \\
Swedish     &  58,547 &  1 \\
\hline
Total     & 1,643,735 & 83 \\
\hline
   \end{tabular}
   \label{tab:tweet_data}
\end{table}

We have engaged 83 native speakers (except for English) to manually annotate 
for sentiment over 1.6 million of the collected tweets. 
The annotation process was supported by the Goldfinch platform 
designed specifically for sentiment annotation of short texts
(such as Twitter posts, Facebook comments, ...). 
The annotators were instructed to label each tweet as either
\textit{negative}, \textit{neutral}, or \textit{positive},
by estimating the emotional attitude of the user who posted the tweet.
They could also skip the inappropriate or irrelevant tweets.
The breakdown of the annotated tweets by language is in Table~\ref{tab:tweet_data}.

Another source of data comes from Emojitracker (\url{http://emojitracker.com/}).
Emojitracker monitors and counts the number of emojis used on Twitter in realtime.
It has been active since July 2013, and so far it has detected over 10 billion emoji occurrences.
We downloaded the current count of emoji occurrences as of June 2015.
This data is used to estimate how representative is our sample of emojis
in the annotated tweets.

The data from both sources is available in a public language-resource 
repository \textsc{clarin.si} at \url{http://hdl.handle.net/11356/1048}.
There are two data tables, in an open csv format, one for the \esr, and the
other from Emojitracker. The tables list all the emojis found, their
occurrences, and, in the case of the \esr, also their numbers in the negative, 
neutral, and positive tweets. From this data, the \esr\, web page at
\url{http://kt.ijs.si/data/Emoji_sentiment_ranking/} is automatically generated.

\subsection{Emoji Unicode symbols}

The exact definition of what constitutes an emoji symbol is still emerging.
In particular, there is some discrepancy between our set of emojis and 
the emojis tracked by Emojitracker.
Also, during the writing of this paper, in August 2015, the Unicode consortium 
published a new set of emojis,
the \textbf{Unicode Emoji Charts} (\url{http://www.unicode.org/emoji/}).

The set of emojis in our \esr\, follows the Unicode standard version 8 \cite{Unicode8}
and consists of all the single-character symbols from the Unicode category 
`Symbol, Other' (abbreviated \So) that appear in our tweets.
Emojitracker, on the other hand, also tracks some double-character symbols
(10 Country Flags, and 11 Combining Enclosing Keycaps), 
but does not track all the \So\, symbols that appear in our data.
In particular, 49 Dingbats, 46 Miscellaneous Symbols, 
38 Box Drawings, 28 Geometric Shapes, 21 Enclosed Alphanumerics, 
20 Enclosed Alphanumeric Supplements, and 13 Arrows are not tracked.
The Unicode Emoji Charts have introduced even more new emoji symbols,
in particular an exhaustive list of 257 double-character Country Flags.
A comparison of the overlaps and differences in the emoji symbol
specifications between the three sources is in
Tables~\ref{tab:emoji_symbols} and~\ref{tab:emoji_data}.


\begin{table}[h]
  \centering
  \caption{\textbf{Types and numbers of emoji symbols.}
  \So\, is an abbreviation for the Unicode category `Symbol, Other'.}
  \begin{tabular}{|l|r|r|r|}
\hline
                            & No. of  & single     & \So \\     \cline{4-4}
                            & all     & character  & non-\So \\ \cline{3-4}
                            & emoji   & double     & flags \\   \cline{4-4}
                            & symbols & character  & keycaps \\
\hline
\hline
\textbf{\esr}               &  969 &  969 & 969 \\ \cline{4-4}
~                           &      &      &   0 \\ \cline{3-4}
~                           &      &    0 &   0 \\ \cline{4-4}
~                           &      &      &   0 \\
\hline
\textbf{Emojitracker}       &  845 &  824 & 812 \\ \cline{4-4}
~                           &      &      &  12 \\ \cline{3-4}
~                           &      &   21 &  10 \\ \cline{4-4}
~                           &      &      &  11 \\
\hline
\textbf{Unicode Emoji Charts} & 1281 & 1012 & 995 \\ \cline{4-4}
~                           &      &      &  17 \\ \cline{3-4}
~                           &      &  269 & 257 \\ \cline{4-4}
~                           &      &      &  12 \\
\hline
   \end{tabular}
   \label{tab:emoji_symbols}
\end{table}

\begin{table}[h]
  \centering
  \caption{\textbf{Overlaps and differences for emojis from the three data sources.}
  A table entry is the number of emojis in ($\in$), or missing ($\notin$) from a data source.
  $N\;(Single, Double)$ denotes the total number $N$ of emoji symbols, partitioned into
  the $Single$- and $Double$-character symbols, respectively.}
  \begin{tabular}{|l|l|c|c|c|}
\hline
\multicolumn{2}{|l|}{~} & \multicolumn{2}{|l|}{\textbf{\esr}} & \\
\multicolumn{2}{|l|}{~} & $\in$ & $\notin$ & Total \\
\hline
\textbf{Emojitracker} & $\in$    & 721 (721, 0) & 124 (103, 21) & 845 (824, 21) \\
                     & $\notin$ & 248 (248, 0)  &   /           & /   \\
\hline
\textbf{Unicode}    & $\in$    & 734 (734, 0) & 547 (278, 269) & 1281 (1012, 269) \\
\textbf{Emoji Charts} & $\notin$ & 235 (235, 0) &   /          & /    \\
\hline
\multicolumn{2}{|l|}{Total} & 969 (969, 0) & / & / \\
\hline
   \end{tabular}
   \label{tab:emoji_data}
\end{table}

The emoji symbols that are not common to all the three data sources are
relatively infrequent. The highest-ranking emoji in Emojitracker, which
is absent from our data, has the rank 157 (double exclamation mark).
The highest-ranking emoji in the \esr, not tracked by Emojitracker,
has the rank 13 (white heart suit).
Additionally, we noticed that we missed three characters from the
\So\, category: `degree sign', `numero sign', and `trade mark sign'.
However, only the `trade mark sign' (with 257 occurrences in our data) is also considered 
by the Emojitracker and the Unicode Emoji Charts.
Despite these minor differences in the emoji sets, all our results remain valid.
However, in the next version of the \esr\, we plan to extend our set to
double-character symbols, and consider all the
emojis from the Unicode Emoji Charts as an authoritative source.

\subsection{Sentiment formalization}

The sentiment of an individual tweet can be 
\textit{negative}, \textit{neutral}, or \textit{positive}.
Formally, we represent it by a discrete, 3-valued variable, $c$,
which denotes the sentiment class:
$$
c \in \{-1, 0, +1\}
$$
This variable models well our assumptions about the ordering of 
the sentiment values and the distances between them.

An object of Twitter posts to which we attribute sentiment (an emoji 
in our case, but it can also be 
a stock \cite{Ranco2015},
a political party \cite{smailovic2015}, 
a discussion topic \cite{Sluban2015, Zollo2015}, etc.)
occurs in several tweets. A discrete distribution:
$$
N(c) \;,\; \sum_{c}N(c) = N \;,\; c \in \{-1, 0, +1\} \;,
$$
captures the sentiment distribution for the set of relevant tweets.
$N$ denotes the number of all the occurrences of the object in the tweets,
and $N(c)$ are the occurrences in tweets with the sentiment label $c$.
From the above we form a discrete probability distribution:
$$
(p_{-}, p_{0}, p_{+}) \;,\; \sum_{c}p_c = 1 \;.
$$
For convenience, we use the following abbreviations:
$$
p_{-} = p(-1) \;,\; p_{0} = p(0) \;,\; p_{+} = p(+1) \;,
$$
where $p_{-}$, $p_{0}$, and $p_{+}$ denote the \textbf{negativity}, 
\textbf{neutrality}, and \textbf{positivity} 
of the object (an emoji in our case), respectively.
In SentiWordNet \cite{baccianella2010sentiwordnet}, 
the term \textbf{objectivity} is used instead of the neutrality $p_{0}$.
The \textbf{subjectivity} can then be defined as
$p_{-} + p_{+}$ \cite{Zhang2010}.

Typically, probabilities are estimated from relative frequencies,
$p_{c} = N(c)/N$. For large samples, such estimates are good approximations.
Often, however, and in particular in our case, we are dealing with small
samples, e.g., $N = 5$.
In such situations, it is better to use the \textit{Laplace estimate} (also
known as the \textit{rule of succession}) to estimate the 
probability \cite{Good1965}:
$$
p_{c} = \frac{N(c) + 1}{N + k} \;,\; 
( \text{for large}\; N: \; p_{c} \approx \frac{N(c)}{N} ) \;.
$$
The constant $k$ in the denominator is the cardinality of the class,
in our case $k = |c| = 3$. 
The Laplace estimate assumes a prior uniform distribution, which makes
sense when the sample size is small.

Once we have a discrete probability distribution, with properly estimated
probabilities, we can compute its mean:
$$
\bar{x} = \sum_{c}p_c\cdot c \;.
$$
We define the \textbf{sentiment score}, $\bar{s}$, as the mean of the
discrete probability distribution:
$$
\bar{s} = -1\cdot p_{-} + 0\cdot p_{0} + 1\cdot p_{+} = p_{+} - p_{-} \;.
$$
The sentiment score has the range: $-1 < \bar{s} < +1$.

The standard deviation of a discrete probability distribution is:
$$
\sd = \sqrt{\sum_{c}p_c\cdot (c - \bar{x})^{2}} \;,
$$
and the standard error of the mean is:
$$
\sm = \frac{\sd}{\sqrt{N}} \;.
$$

\subsection{Sentiment bar}

\begin{figure}[h]
\begin{center}
\includegraphics[width=13cm]{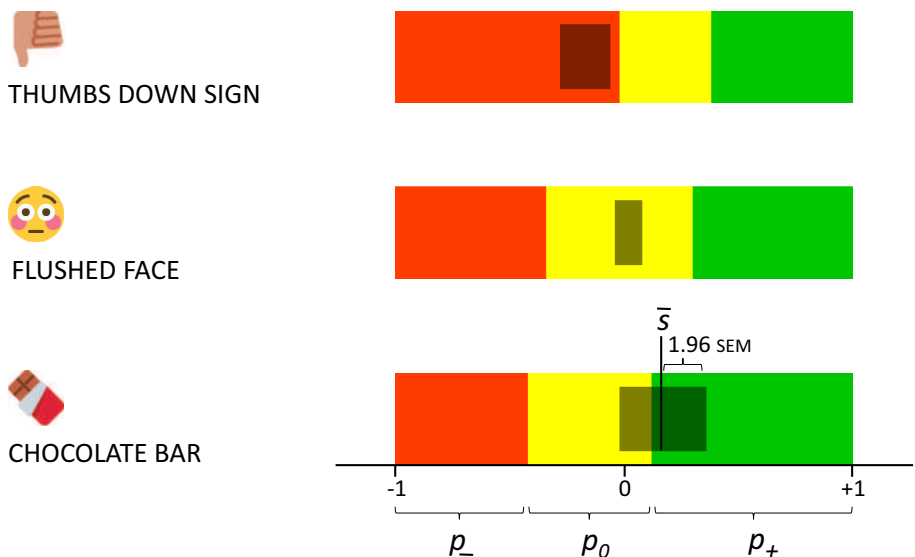}
\caption{\textbf{Sentiment bars of the `thumbs down sign', `flushed face', and `chocolate bar' emojis.} 
The colored bar extends from $-1$ to $+1$, the range of the sentiment score $\bar{s}$.
The grey bar is centered at $\bar{s}$ and extended for $\pm 1.96 \sm$, 
but never beyond the range of $\bar{s}$.
Colored parts are proportional to negativity ($p_{-}$, red), 
neutrality ($p_{0}$, yellow), and positivity ($p_{+}$, green).}
\label{fig:sent_bar}
\end{center}
\end{figure}

The sentiment bar is a useful, novel visualization of the sentiment 
attributed to an emoji (see 
\url{http://kt.ijs.si/data/Emoji_sentiment_ranking/} for examples).
In a single image, it captures all the sentiment properties, computed from the
sentiment distribution of the emoji occurrences:
$p_{-}, \; p_{0}, \; p_{+}, \; \bar{s}$, and $\bar{s} \pm 1.96 \sm$
(the 95\% confidence interval).
Three examples that illustrate how the sentiment properties are
mapped into the graphical features are shown in Fig~\ref{fig:sent_bar}.
The top sentiment bar corresponds to the `thumbs down sign' emoji,
and indicates negative sentiment, with high confidence.
The middle bar represents the estimated sentiment of the `flushed face' emoji.
The sentiment is neutral, close to zero, where both negative and 
positive sentiment are balanced.
The bottom bar corresponds to the `chocolate bar' emoji.
Its sentiment score is positive, but its standard error bar extends into
the neutral zone, so we can conclude with high confidence only that its sentiment
is not negative.

\subsection{Welch's t-test}

Welch's t-test \cite{Welch1947} is used to test the hypothesis that 
two populations have equal means. It is an adaptation of Student's
t-test, but is more reliable when the two samples have unequal 
variances and sample sizes. Welch's t-test is also robust for 
skewed distributions and even more for large sample sizes \cite{Fagerland2012}.

Welch's t-test defines the t statistic as follows:
$$
t = \frac{\bar{x}_{1} - \bar{x}_{2}}{\sqrt{\frac{\sd_1^2}{N_1} + \frac{\sd_2^2}{N_2}}} \,.
$$

The degrees of freedom, $\nu$, are estimated as follows:
$$
\nu \approx \left\lfloor \frac{(\frac{\sd_1^2}{N_1} + \frac{\sd_2^2}{N_2})^2}{\frac{\sd_{1}^{4}}{N_{1}^{2} (N_{1}-1)} + \frac{\sd_{2}^{4}}{N_{2}^{2} (N_{2}-1)}} \right\rfloor \,,
$$
where $\lfloor \, \rfloor$ denotes the approximate degrees of freedom,
rounded down to the nearest integer.
Once the t value and the degrees of freedom are determined, 
a p-value can be found from a table of values for Student's t-distribution. 
For large degrees of freedom, $\nu > 100$, the t-distribution is 
very close to the normal distribution.
If the p-value is below the threshold of statistical significance,
then the null hypothesis is rejected.

\subsection{Pearson and Spearman correlations}

We need to correlate two properties of the \esr\, with other data.
In the first case we correlate the emojis ranked by occurrence
to the Emojitracker list---the property of the list elements
is the number of occurrences.
In the second case we correlate the emojis ranked by sentiment
to subsets of emojis from the 13 different languages---the property of
the list elements is the sentiment score.

For any two lists $x$ and $y$, of length $n$, we first compute the 
Pearson correlation coefficient~\cite{Pearson1895}:
$$
r(x,y) = \frac{\sum_{i=1}^{n}{(x_{i}-\bar{x})(y_{i}-\bar{y})}}{\sqrt{\sum_{i=1}^{n}{(x_{i}-\bar{x})^2\sum_{i=1}^{n}{(y_{i}-\bar{y})^2}}}} \;,
$$
where $\bar{x}$ and $\bar{y}$ are the list's mean values, respectively.
The Spearman's rank correlation coefficient~\cite{Spearman1904} is
computed in the same way, the property values of the $x$ and $y$
elements are just replaced with their ranks.
In both cases we report the
correlation coefficients at the 1\% significance level.

\subsection{Agreement measures}

In general, an agreement can be estimated between any two methods
for generating data. In our case we want to estimate the agreement between
humans when annotating the same tweets for sentiment.
A comparison of agreements between different datasets gives some
clue about how difficult the task is. There are different measures of
agreement, and to get a robust estimate of the differences, we
apply three well-known measures.

Krippendorff's \alfa-reliability \cite{Krippendorff2012} is a generalization of several
specialized agreement measures. It works for any number of annotators,
is applicable to different variable types and metrics 
(e.g., nominal, ordered, interval, etc.), and can handle small sample sizes.
\alfa\, is defined as follows:
$$
\mathit{Alpha} = 1 - \frac{D_{o}}{D_{e}} \,,
$$
where $D_{o}$ is the observed disagreement between the annotators, and
$D_{e}$ is the disagreement expected by chance.
When the annotators agree perfectly, \alfa\;$=1$, and when the level of 
agreement equals the agreement by chance, \alfa\;$=0$. 
The two disagreement measures are defined as follows:
$$
D_{o} = \frac{1}{N} \sum_{c,c'} N(c,c') \cdot \delta^2(c,c') \,,
$$
$$
D_{e} = \frac{1}{N(N-1)} \sum_{c,c'} N(c) \cdot N(c') \cdot \delta^2(c,c') \,.
$$
The arguments, $N, N(c,c'), N(c)$, and $N(c')$,
refer to the frequencies in a coincidence matrix, defined below.
$\delta(c,c')$ is a difference function between the values of $c$ and $c'$,
and depends on the metric properties of the variable.
In our case, for the discrete sentiment variables $c$ and $c'$,
the difference function $\delta$ is defined as:
$$
\delta(c,c') = |c - c'| \;\;\;\; c,c'\in \{-1,0,+1\} \;.
$$
In \cite{Krippendorff2012}, this is called the \textit{interval} difference function.
Note that the function attributes a disagreement of $1$ between the
\textit{negative} (or \textit{positive}) and the \textit{neutral} sentiment,
and a disagreement of $2$ between the \textit{negative} and \textit{positive}
sentiments.

A \textbf{coincidence matrix} tabulates all the pairable values of $c$ from two
different annotators into a $k$-by-$k$ square matrix, where $k = |c|$.
Unlike a contingency matrix (used in association and correlation statistics)
which tabulates pairs of values, a coincidence matrix tabulates all the pairable values.
A coincidence matrix omits references to annotators.
It is symmetrical around the diagonal, which contains all the perfect matches.
A coincidence matrix has the following general form:
$$
\begin{array}{c|ccc|c}
  &   & c' &   & \sum \\
\hline
  & . & . & . & \\
c & . & N(c,c') & . & N(c) \\
  & . & . & . & \\
\hline
\sum  &   & N(c') & & N \\
\end{array}
$$
Here $c$ and $c'$ range over all possible values of the variable.
In a coincidence matrix, each labeled unit is entered twice,
once as a $(c,c')$ pair, and once as a $(c',c)$ pair.
$N(c,c')$ is the number of units labeled by the values $c$ and $c'$
by different annotators, $N(c)$ and $N(c')$ are the totals
for each value, and $N$ is the grand total.
Note that $N$ is two times the number of units labeled by
the different annotators.

In the case of sentiment annotations, we have a
$3$-by-$3$ coincidence matrix.
Two example matrices are shown in 
Tables~\ref{tab:inter_emojis} and \ref{tab:inter_noemojis}.
Note that both coincidence matrices in 
Tables~\ref{tab:inter_emojis} and \ref{tab:inter_noemojis} are
symmetric around the diagonal, 
and that the totals $N$ are two times larger
than in Table~\ref{tab:inter_annotator_agreement} because each
annotated tweet is counted twice.

\begin{table}[h]
    \centering
    \caption{ \textbf{Coincidence matrix for tweets with emojis.}}
    \begin{tabular}{|l|rrr|r|}
\hline
Sentiment & \textit{Negative} & \textit{Neutral}  & \textit{Positive} & Total \\
\hline
\textit{Negative} & 1,070 & 354 & 196 & 1,620 \\
\textit{Neutral}  & 354 & 902 & 725 & 1,981 \\
\textit{Positive} & 196 & 725 & 2,572 & 3,493 \\
\hline
Total             & 1,620 & 1,981 & 3,493 & 7,094 \\
\hline
    \end{tabular}
\label{tab:inter_emojis}
\end{table}

\begin{table}[h]
    \centering
    \caption{ \textbf{Coincidence matrix for tweets without emojis.}}
    \begin{tabular}{|l|rrr|r|}
\hline
Sentiment & \textit{Negative} & \textit{Neutral}  & \textit{Positive} & Total \\
\hline
\textit{Negative} & 15,356 & 7,777 & 3,004 & 26,137 \\
\textit{Neutral}  & 7,777 & 23,670 & 10,921 & 42,368 \\
\textit{Positive} & 3,004 & 10,921 & 21,624 & 35,549 \\
\hline
Total             & 26,137 & 42,368 & 35,549 & 104,054 \\
\hline
    \end{tabular}
\label{tab:inter_noemojis}
\end{table}

In machine learning, a classification model is automatically constructed
from the training data and evaluated on a disjoint test data.
A common, and the simplest, measure of the performance of the model is \acc,
which measures the agreement between the model and the test data.
Here, we use the same measure to estimate the agreement between
the pairs of annotators.
\acc\, is defined in terms of the observed disagreement $D_{o}$:
$$
\mathit{Accuracy} = 1 - D_o = \frac{1}{N} \sum_{c} N(c,c) \,.
$$
\acc\, is simply the fraction of the diagonal elements of the
coincidence matrix. Note that it does not account for the
(dis)agreement by chance, nor for the ordering between the sentiment values.

Another, more sophisticated measure of performance, specifically
designed for 3-class sentiment classifiers \cite{kiritchenko2014sentiment},
is \favg($-,+$):
$$
\overline{F_1}(-,+) = \frac{F_1(-) + F_1(+)}{2} \,.
$$
\favg($-,+$) implicitly takes into account the ordering of the sentiment values
by considering only the \textit{negative} $(-)$ and \textit{positive} $(+)$
labels, and ignoring the middle, \textit{neutral} label.
In general, $F_{1}(c)$ (known as the F-score) is a harmonic mean of
precision and recall for class $c$. In the case of a coincidence matrix,
which is symmetric, the `precision' and `recall' are equal, 
and thus $F_{1}(c)$ degenerates into:
$$
F_{1}(c) = \frac{N(c,c)}{N(c)} \,.
$$
In terms of the annotator agreement, $F_{1}(c)$ is the fraction of
equally labeled tweets out of all the tweets with label $c$.

\section*{Acknowledgments}

This work was supported in part by the EC projects SIMPOL (no. 610704), 
MULTIPLEX (no. 317532) and DOLFINS (no. 640772), 
and by the Slovenian ARRS programme Knowledge Technologies (no. P2-103).

We acknowledge Gama System (\url{http://www.gama-system.si})
who collected most of the tweets (except English), and
Sowa Labs (\url{http://www.sowalabs.com}) for providing 
the Goldfinch platform for the sentiment annotation of the tweets.
We thank Sa\v{s}o Rutar for generating the \esr\, web page,
Andrej Blejec for statistical insights, and
Vinko Zlati\'{c} for suggesting an emoji distribution model.


%
%
%


\end{document}